\pdfoutput=1

\documentclass[11pt]{article}

\usepackage[]{emnlp2021}
\usepackage{graphicx}

\usepackage{times}
\usepackage{latexsym}

\usepackage[T1]{fontenc}

\usepackage[utf8]{inputenc}

\usepackage{microtype}

\usepackage{amsmath}

\usepackage{multicol}
\usepackage[linesnumbered,ruled,vlined,algo2e]{algorithm2e}
\usepackage{setspace}

\usepackage[utf8]{inputenc} 
\usepackage[T1]{fontenc}    
\usepackage{hyperref}       
\usepackage{url}            
\usepackage{booktabs,multirow}       
\usepackage{amsfonts}       
\usepackage{nicefrac}       
\usepackage{microtype}      

\usepackage{graphicx}
\usepackage{times}
\usepackage{epsfig}
\usepackage{graphicx}
\usepackage{amsmath}
\usepackage{amssymb}
\usepackage{xcolor,vwcol}

\title{Caption Enriched Samples for Improving Hateful Memes Detection}

\author{Efrat Blaier\thanks{\;\; Denotes equal contribution.} \\
  Tel-Aviv University\\\And
  Itzik Malkiel$^{*}$ \\
  Tel-Aviv University \\\And
  Lior Wolf \\
  Tel-Aviv University
  }

\begin{document}
\maketitle
\begin{abstract}
The recently introduced hateful meme challenge demonstrates the difficulty of determining whether a meme is hateful or not. Specifically, both unimodal language models and multimodal vision-language models cannot reach the human level of performance. Motivated by the need to model the contrast between the image content and the overlayed text, we suggest applying an off-the-shelf image captioning tool in order to capture the first. We demonstrate that the incorporation of such automatic captions during fine-tuning improves the results for various unimodal and multimodal models. Moreover, in the unimodal case, continuing the pre-training of language models on augmented and original caption pairs, is highly beneficial to the classification accuracy. Our code is publicly available \footnote{https://github.com/efrat-safanov/caption-enriched-samples-research}.

\end{abstract}

\section{Introduction}

Multimodal transformers, including VisualBERT~\cite{li2019VisualBERT}, VilBERT~\cite{NEURIPS2019_c74d97b0}, and UNITER~\cite{chen2020uniter} are currently the state of the art methods in tasks that involve both text and images, such as visual question answering~\cite{antol2015vqa} and image captioning~\cite{chen2015microsoft}. One particular example is that of the ``hateful memes'' challenge~\cite{Kiela2020TheHM}, in which the task is to classify whether a given meme is hateful or not. 
By construction (and similar to real-world memes), the classifier has to apply complex reasoning that involves a multimodal text-image analysis from both image and text. The prevalence of sarcasm adds another layer of difficulty to both machines and humans. 
\citet{Kiela2020TheHM} results indicate that the  human accuracy is only about 85\%. Recent state-of-the-art models perform considerably less accurately, achieving up to 64.73\% accuracy. 

In this work, we adopt an off-the-shelf caption generator, and show that combining such captions as a third input (in addition to the image and the overlayed text) can effectively improve the performance of both multimodal and unimodal models.  {Two techniques are used: continued pre-training of the baseline model with the additional augmented captions, and using the added information as part of the fine-tuned model used for the hateful memes task.} 

\begin{figure}[t]
\centering
\includegraphics[width=0.49\textwidth]{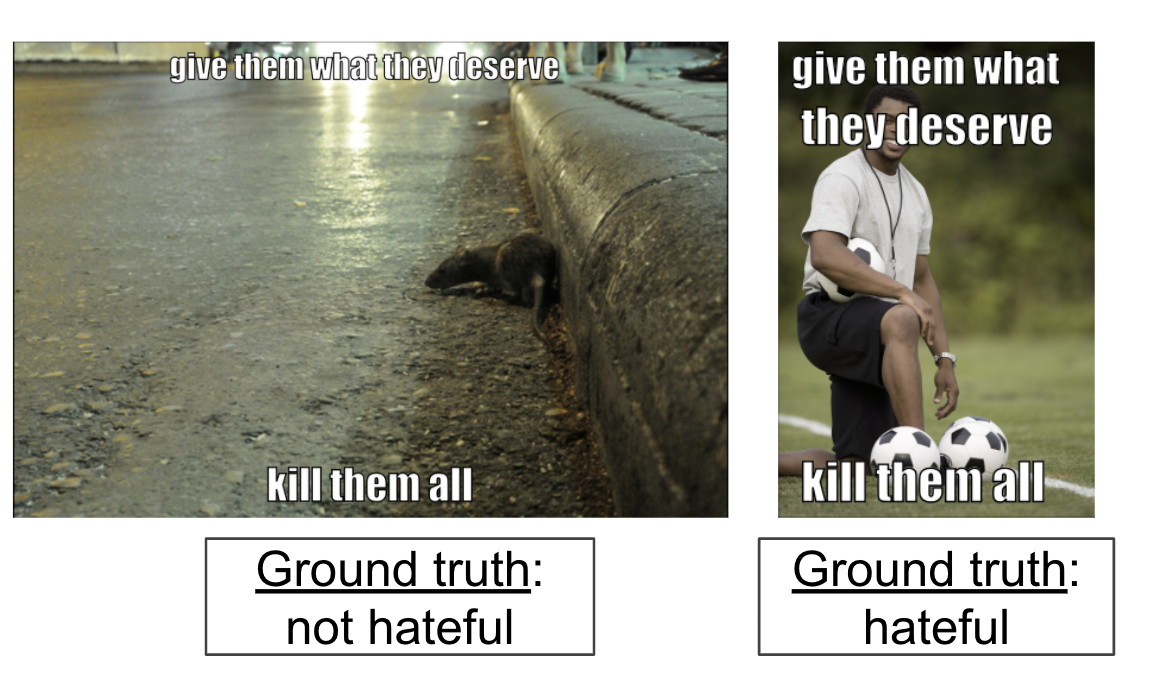}

\caption{An example of memes in which the classifiction depends on both the text and the image. Specifically, in this {\em Benign Confounder} example, the same text appears and the classification changes depending on the image.}
\label{fig:benign}
\end{figure}

\section{Related Work}

Transformer-based Language Models (LMs) have shown significant performance gains in various natural language understanding tasks, such as text-similarity and language inference \cite{devlin2019bert, liu2019roberta}. These models are usually pre-trained on the Masked Language Model (MLM) objective followed by a task-specific fine-tuning process \cite{wang-etal-2018-glue}. 

Recently, vision and language tasks have gained a lot of traction, where transformer-based multimodality models were introduced, showing great promise in solving multimodality problems, such as visual reasoning \cite{suhr2019corpus}, image caption generation \cite{chen2015microsoft} and more. 

In \citet{Kiela2020TheHM}, the authors introduced the Hateful Memes challenge - a dataset of 10,000 memes, each associated with an image and text. Each meme sample image-text pair is binary labeled as hateful or not hateful. The dataset is split into 8,500 train memes, 500 and 1,000 validation and test samples, respectively. The test labels are not public and are used for leaderboard scoring of models.

The dataset was artificially constructed based on real-world memes. To ensure that the dataset requires multimodal reasoning, image and text confounders were crafted to enrich the dataset with contrastive examples. Confounders samples are memes that appear in two variations: same image with different texts that have different labels or vice versa (see Fig.~\ref{fig:benign}). 

In \citet{Kiela2020TheHM}, the authors show that both unimodal models (such as Bert \cite{devlin2019bert}) and multimodal models (e.g. VisualBERT \cite{li2019VisualBERT}) seem to produce relatively poor performance, that is significantly inferior to the human accuracy (reported to be 84.7\%). This can be attributed to the task difficulty, which requires an advanced reasoning process, where some samples are found to also confuse humans. 

The task of caption generation has been shown to yield promising results in various works, such as the show and tell \cite{vinyals2015tell}, show tell and attend \cite{xu2016show}, and more. Among the different caption generator techniques, the Neural Image Caption (NIC) \cite{Vinyals_2017} has been shown to yield promising results on the various caption tasks. The NIC model is an encoder-decoder network, trained to generate captions from images. We use this generator to augment the hateful memes dataset, by associating each meme image-text pair with an additional auto-generated caption. In our work, we adopt the implementation of the IBM Image Caption Generator\footnote{github.com/IBM/MAX-Image-Caption-Generator}. 

\citet{gururangan2020dont} show that continued pre-training of a pre-trained language model on a given dataset at hand improves the performance of the fine-tuning procedure. In a similar spirit, we demonstrate that for the hateful meme task, it can be beneficial to continue pre-training on text-pairs of meme text and augmented captions. 

\begin{figure*}[t]
\centering
\includegraphics[width=1.0\textwidth]{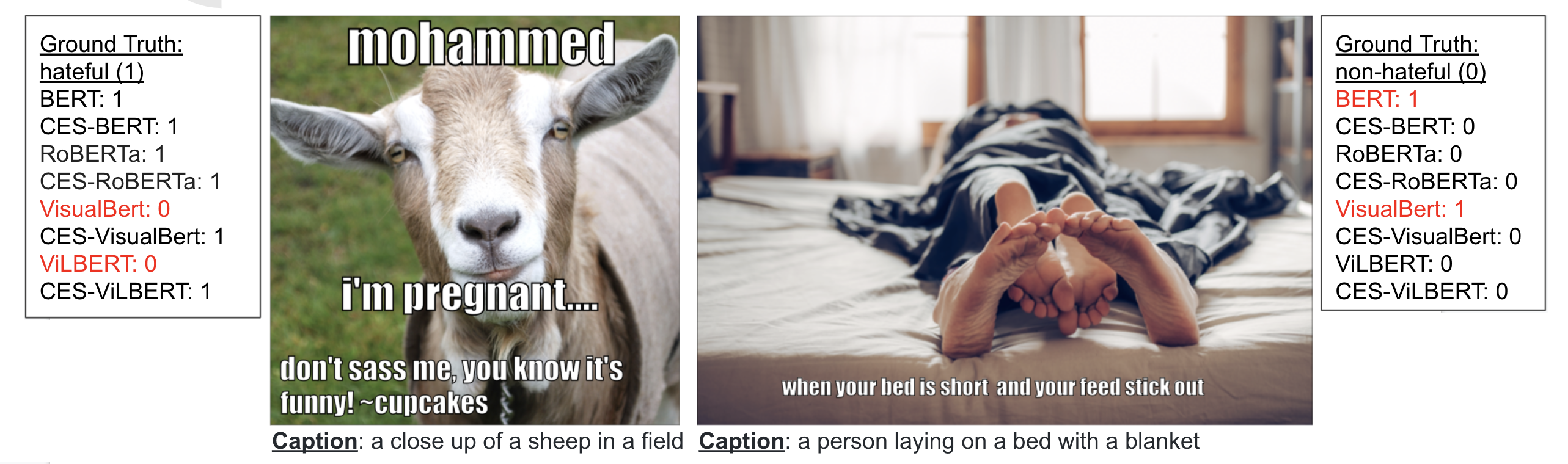}
\caption{Two representative samples of unimodal and multimodal predictions, with and without the CES approach.}
\label{fig:captions_better_VisualBERT}
\end{figure*}

\section{Method}

The hateful memes challenge is a binary classification problem in which each meme sample $x=(m,c)$ consists of an image in some image domain $M$ and an associated text in a language domain $C$. In both cases, we employ an off-the-shelf image-caption generator $G: M \rightarrow C$, which receives an image from domain $M$ and returns a caption from $C$. Every sample thus becomes $x^{*} = (m,c,c^{*})$, where $c^{*}=G(m)$.

We next present the training strategy, termed Caption Enriched Samples (CES), in either the unimodal or the multimodal settings. In the former, the network classifier considers only text input, and in the latter image-text inputs.

\begin{table}[t]
\centering
\resizebox{1.0\linewidth}{!}{
\begin{tabular}{lcccc}
\toprule
&\multicolumn{2}{c}{\textbf{Validation}} & \multicolumn{2}{c}{\textbf{Test}}\\
\cmidrule(lr){2-3}
\cmidrule(lr){4-5}
  & \textbf{Baseline} & \textbf{CES}
  & \textbf{Baseline} & \textbf{CES}\\
\hline
 Human &  -&  & 82.65 & -\\
BERT & 64.65 & \textbf{69.32}  & 65.08 & \textbf{70.70} \\ 
RoBERTa & 65.77 & \textbf{72.95} & 65.77 & \textbf{72.37} \\
VisualBERT & 73.97 & \textbf{75.55}  & 71.41 & \textbf{72.93} \\
ViLBERT  & 70.07 & \textbf{75.98} &  70.03 & \textbf{72.59} \\

\bottomrule
\end{tabular}}
\caption{AUROC for various models reported on validation and test sets of the hateful memes dataset.}
\label{tab:results}
\end{table}

\subsection{The unimodal case}

In unimodal settings, we employ a dual-phase training approach, which utilizes caption-pairs composed of the original caption and the generated caption $(c,c^*)$. In the first training phase, we consider the caption-pairs of each sample, tokenize each element, concatenate them using the special [SEP] token, and apply a standard masking procedure. Similar to BERT, the training objective is to optimize a standard masked language model (MLM). Notably, in contrast to \cite{gururangan2020dont}, our method applies this pre-training on an augmented text, where half of the data was generated by an image-to-text generator.

In the second phase, we fine-tune the same model on the task at hand, by feeding the model with the pairs $(c,c^*)$. During this phase, we employ a standard fine-tuning approach, where we initialize a classification head on top of the pre-trained model and minimize a binary cross-entropy loss.

\subsection{The multimodal case}
In multimodal settings, in order to reduce training costs, we do not continue the pre-training phase of the multimodal backbones on the triplets $(m,c,c^*)$, as done in the unimodal case, and employ only a fine-tuning strategy. 

We, therefore, adopt a multimodal backbone {network that was pre-trained on a general vision-language task}, and initialize a classification head composed of two fully connected layers on top. We fine-tune the models, by propagating the triplets $(m,c,c^*)$ and optimizing a standard binary cross entropy loss for the prediction of the meme labels (hateful/non hateful). 

The classification head is employed over the pooled representation of the embeddings. The pooled embeddings are the ones proposed by each of the multimodal models at hand, as detailed in Sec.\ref{label:results}. For example, in the ViLBERT model~\cite{NEURIPS2019_c74d97b0}, the learned classification head is employed on top of the element-wise product of the image and text representations, which are the embeddings of the first tokens obtained from each of the image sequence and the textual sequence. Note that when employing CES on the ViLBERT, the textual sequence is a pair of sentences separated by the special [SEP].

\begin{table}[t]
\centering
\resizebox{1.0\linewidth}{!}{%
\begin{tabular}{lcc}
\toprule
\textbf{Model} & \textbf{Validation} & \textbf{Test}\\
\midrule
BERT & 64.65 & 65.08 \\
i CES-BERT
& 69.32 & 70.7 \\
ii CES-BERT 
& 65.19 & * \\
iii CES-BERT
& 69.25 & * \\
CES-BERT  & \textbf{69.76} & * \\
 \midrule
RoBERTa & 65.44 & 65.77\\

i CES-RoBERTa 
& {70.47} & {70.91} \\
ii CES-RoBERTa
& 64.8 & * \\
iii CES-RoBERTa
& 71.11 & 71.08 \\
CES-RoBERTa & \textbf{72.95} & \textbf{72.37} \\
\midrule
UNITER &77.01 & 78.68\\
UNITER+RoBERTa & 78.04 & *\\
UNITER+i.CES-RoBERTa & 77.53 & *\\
UNITER+ii.CES-RoBERTa & \textbf{78.31}  & *\\
UNITER+iii.CES-RoBERTa & \textbf{78.57}  & *\\
UNITER+CES-RoBERTa& \textbf{78.29} & \textbf{78.90}\\
 \bottomrule
\end{tabular}}
\caption{Ablation study results for the unimodal models. Reported are the AUROC scores on the validation and test sets. (*) For some ablations, we do not have the test results, since the Hateful Memes challenge server went offline on May 1st, 2021. }
\label{tab:ablation}
\end{table}

\section{Results}
\label{label:results}
In this section, we evaluate and report the performance of CES applied to the following models:
\paragraph{BERT and RoBERTa} are the BERT$_{\text{BASE}}$ \cite{devlin2019bert} and RoBERTa$_{\text{BASE}}$\cite{liu2019roberta} models, fine-tuned on the meme text using a standard fine-tuning approach.

\paragraph{ViLBERT} is a multimodal Transformer with a two-stream architecture that embeds image regions and language separately, employed with co-attentional transformer layers that allow both streams to interact with each other~\cite{NEURIPS2019_c74d97b0}.

\paragraph{VisualBERT} is another multimodal Transformer that was pre-trained on image-text inputs, by aligning elements of the text and regions in the image through cross-attention and self-attention operations~\cite{li2019VisualBERT}. In contrast to ViLBERT, this model employs a single-stream
architecture, embedding image regions and language tokens through the same transformer blocks.

For the image-caption generator $G$ we employ the IBM MaxCaption model \citet{Vinyals_2017} for generating captions for all memes in the dataset. The length of the generated captions is at most 15 words (see Fig.\ref{fig:captions_better_VisualBERT}).

In Tab.\ref{tab:results}, we report the performance of the above models applied with and without our CES strategy. As can be seen, CES is shown to improve the performance of all models by a sizeable margin. Specifically, for the BERT and RoBERTa unimodal models, we observe that CES improves the test performance by 8.6\% and 10\%, respectively. This can be attributed to the fact that in CES, the unimodal receives a textual representation of the image (the generated caption $c^*$), which is a proxy to the image itself. In the multimodal settings, we observe that CES improves the performance of ViLBERT and VisualBERT by 2.1\% and 3.6\%, respectively.
A similar table reporting the accuracy scores can be found in the appendix. 

Fig.~\ref{fig:captions_better_VisualBERT} depicts two representative samples demonstrating the effectiveness of CES applied to the VisualBERT model. As can be seen in the figure,  the CES-VisualBERT model was able to correctly predict the right label for both images, while the baseline VisualBERT model was confused. 
\subsection{Ablation Study}
We perform an ablation study for the CES technique, by applying multiple variants of it to the BERT and RoBERTa models. We consider three variants: (i) skipping the first phase of pre-training on the pairs $(c, c^*)$ of meme text and generated captions. (ii) applying the continued pre-training followed by fine-tuning, without utilizing the generated captions in both phases (i.e. we solely use the original meme text). (iii) applying continued pre-training solely on the original memes text (without the augmented captions), followed by fine-tuning that utilizes the pairs $(c, c^*)$.

The results, shown in Tab.~\ref{tab:ablation}, indicate that it is important to continue the pre-training on the text-pairs of original meme text and augmented captions, and that the use of the augmented captions during fine-tuning improves the accuracy of the models. 

To further demonstrate the importance of each of the CES components, we evaluate the performance of each RoBERTa variant in an ensemble that includes the state-of-the-art UNITER model. The UNITER model is multi-modal Transformer-based network employing a single-stream architecture~\cite{chen2020uniter} for both vision and textual input. The UNITER pre-training utilizes a combination of four tasks: Masked Language Modeling, Masked Region Modeling, Image-Text Matching (ITM), and Word-Region Alignment (WRA). 

We enhance the UNITER model by either of the RoBERTa ablation variants, by utilizing the late fusion technique form~\cite{Kiela2020TheHM}. Specifically, we concatenate the CLS embeddings of the UNITER and each RoBERTa model, 
and train a two-layer MLP on top. {For the UNITER model, we use the Vilio \citet{muennighoff2020vilio} implementation. }
As can be seen in Tab.~\ref{tab:ablation}, all ensemble variants improve the UNITER performance, while the ones that include a continued pre-trained RoBERTa (UNITER+ii.CES-RoBERTa, UNITER+iii.CES-RoBERTa, UNITER+CES-RoBERTa) yield the largest gains. Following a t-test on the validation set, we observe that those three ensembles yield significant improvement compared to the UNITER and UNITER+RoBERTa baselines (p-value<0.02), while the differences between them are not statistically significant.

\begin{figure}[t]
\centering
\includegraphics[width=0.45\textwidth]{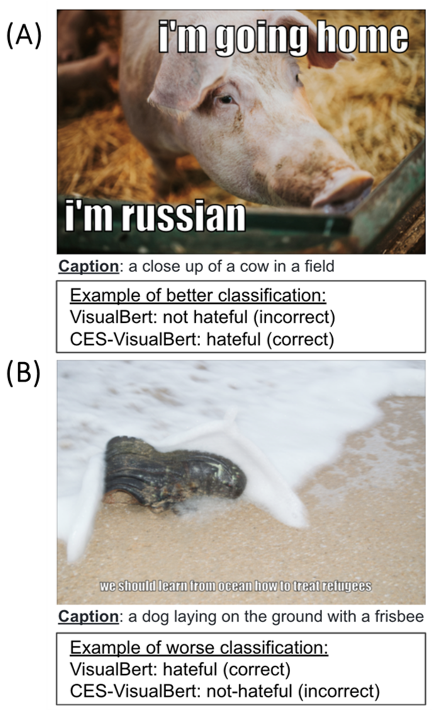}
\caption{Limitations of augmented captions. (A) A representative sample of an inaccurate augmented caption (wrong animal), for which CES was able to recover and improve over the baseline. (B) A representative sample of a wrong caption (detecting a dog playing instead of a shoe), for which CES failed to classify the meme as hateful.}
\label{fig:limitations}
\end{figure}

\subsection{Limitations and Discussion}
CES relies on augmented captions, and, therefore, it also depends on the performance of the caption generator. 
In Fig.~\ref{fig:limitations}(A), we observe a failure case of the caption generator, where the pig in the image was interpreted as a cow. In this case, we observe that CES could recover from the non-accurate caption, perhaps since the cow could be negatively interpreted in this context. 
Fig.~\ref{fig:limitations}(B) demonstrates a case where a wrong augmented caption can confuse CES. In the figure, we observe that the caption generator describes the image as a dog with a frisbee, instead of a shoe getting out of the ocean. As can be seen, CES misclassified the image as not hateful. 

\section{Summary}

{Detecting hateful memes is a challenging task that requires complex image-text reasoning. In this work, we introduce CES, a training approach for both multimodal and unimodal models that leverages an off-the-shelf image-caption generator to enhance the accuracy of hateful meme classifiers. Our results indicate that augmented captions are highly beneficial for transformer-based models, which seem to be able to effectively attend between the augmented caption, meme text, and image.}

\section*{Acknowledgment}
This project has received funding from the European Research Council (ERC) under the European Union Horizon 2020 research and innovation programme (grant ERC CoG 725974). The contribution of the second author is part of a PhD thesis research conducted at Tel Aviv University.

\bibliography{memes}
\bibliographystyle{acl_natbib}

\cleardoublepage
\begin{center}
{\LARGE Supplementary Appendices\footnote{Put here for the reader's convenience.}}
\end{center}

\appendix

\begin{table*}[t]
\centering
\resizebox{1\linewidth}{!}{%
\begin{tabular}{lccccc}
\hline
\textbf{Baseline Source} & \textbf{Model} & \multicolumn{2}{c}{\textbf{Validation}}  & \multicolumn{2}{c}{\textbf{Test}} \\
 &  & \textbf{Baseline} & \textbf{Captions} & \textbf{Baseline} & \textbf{Captions} \\
 &  & \textbf{Acc} & \textbf{Acc} & \textbf{Acc} & \textbf{Acc}\\
\hline
Hateful Memes & human &  & & 84.7 &  \\
  & Bert & 58.8 & \textbf{63.4} & 59.2 & \textbf{63.2} \\
  & Roberta & 59.8 & \textbf{62.2} & 60.8 & \textbf{62.7} \\
& VisualBERT COCO & 65.06 & \textbf{67.8} & 61.7 & \textbf{64.44}\\
& ViLBERT CC & 61.4 & \textbf{65.6} & 61.1 & \textbf{64.4}\\
\hline
\end{tabular}}
\caption{Accuracy of captions usage for various models}
\label{tab:accuracy_table}
\end{table*}

\label{sec:appendix}

\section{Additional Results}

Tab.~\ref{tab:accuracy_table} depicts the accuracy of all models reported in Table 1 from the main text.

\begin{figure*}[t]
\centering
\includegraphics[width=0.8\textwidth]{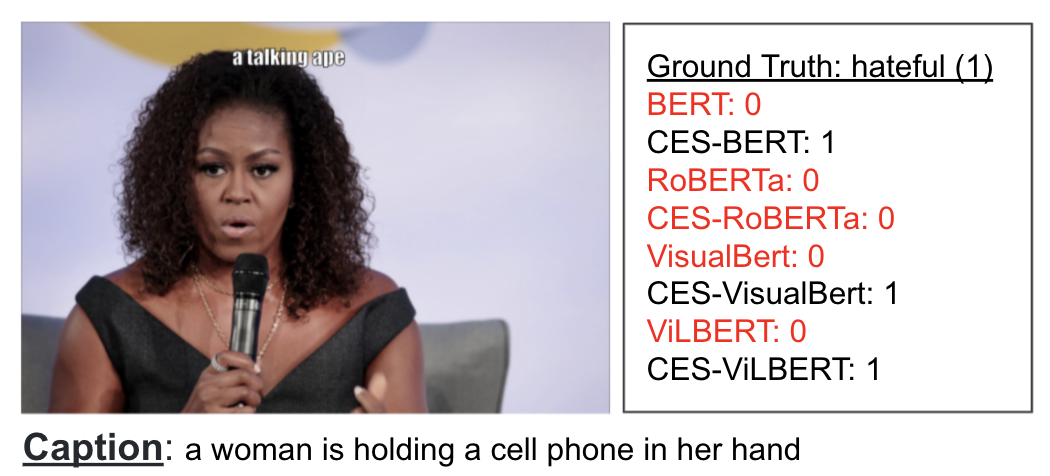}
\caption{An additional sample of inaccurate caption (cellphone vs microphone), for which CES is able to produce better classification in three out of four models}
\label{fig:captions_better_inaccurate}
\end{figure*}

Fig.\ref{fig:captions_better_inaccurate} presents an additional sample with better classification in most CES models, although the generated caption inaccurate.

\begin{figure*}[t]
\centering
\includegraphics[width=0.8\textwidth]{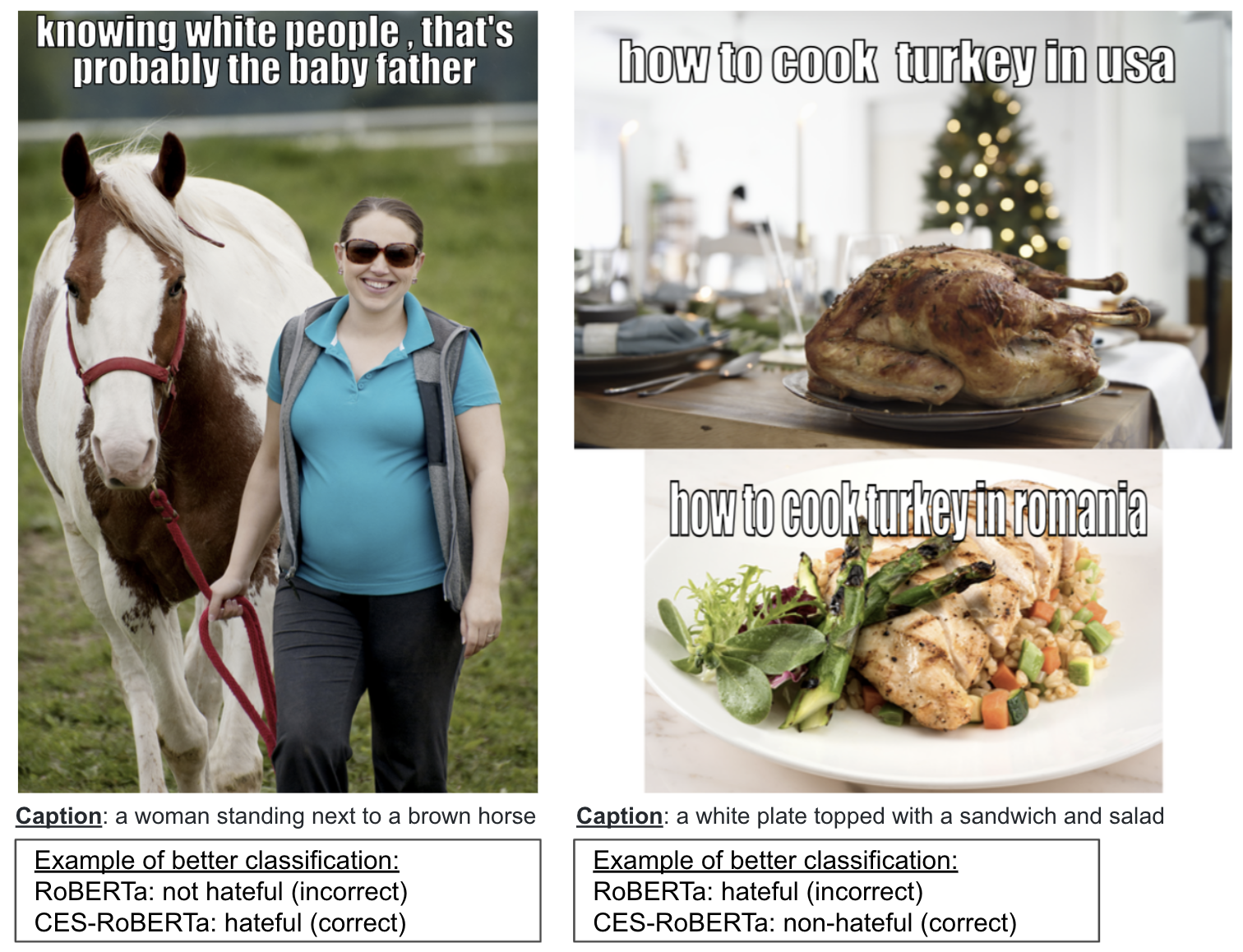}
\caption{RoBERTa results.}
\label{fig:supp_roberta_better}
\end{figure*}

\begin{figure*}[t]
\centering
\includegraphics[width=0.8\textwidth]{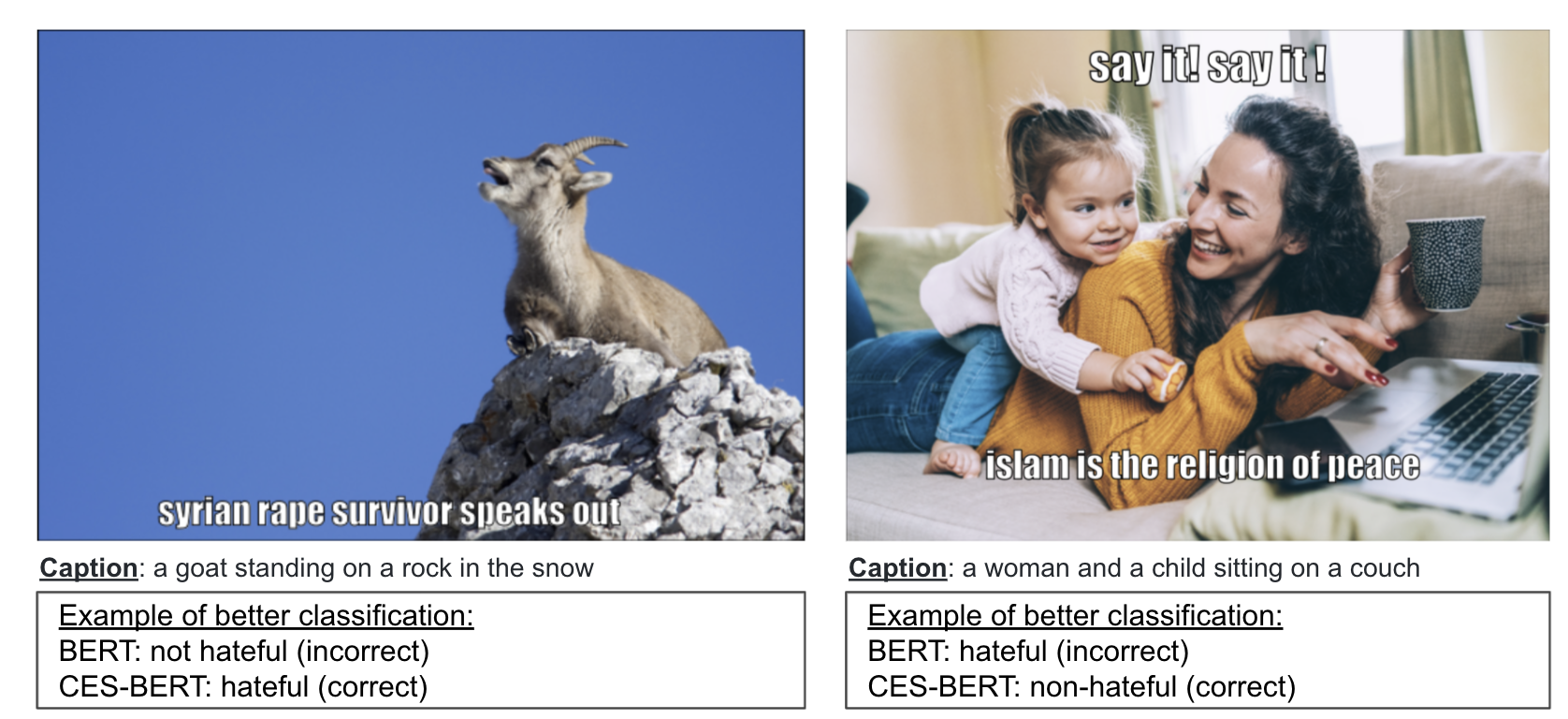}
\caption{BERT results.}
\label{fig:supp_bert_better}
\end{figure*}

\begin{figure*}[t]
\centering
\includegraphics[width=0.8\textwidth]{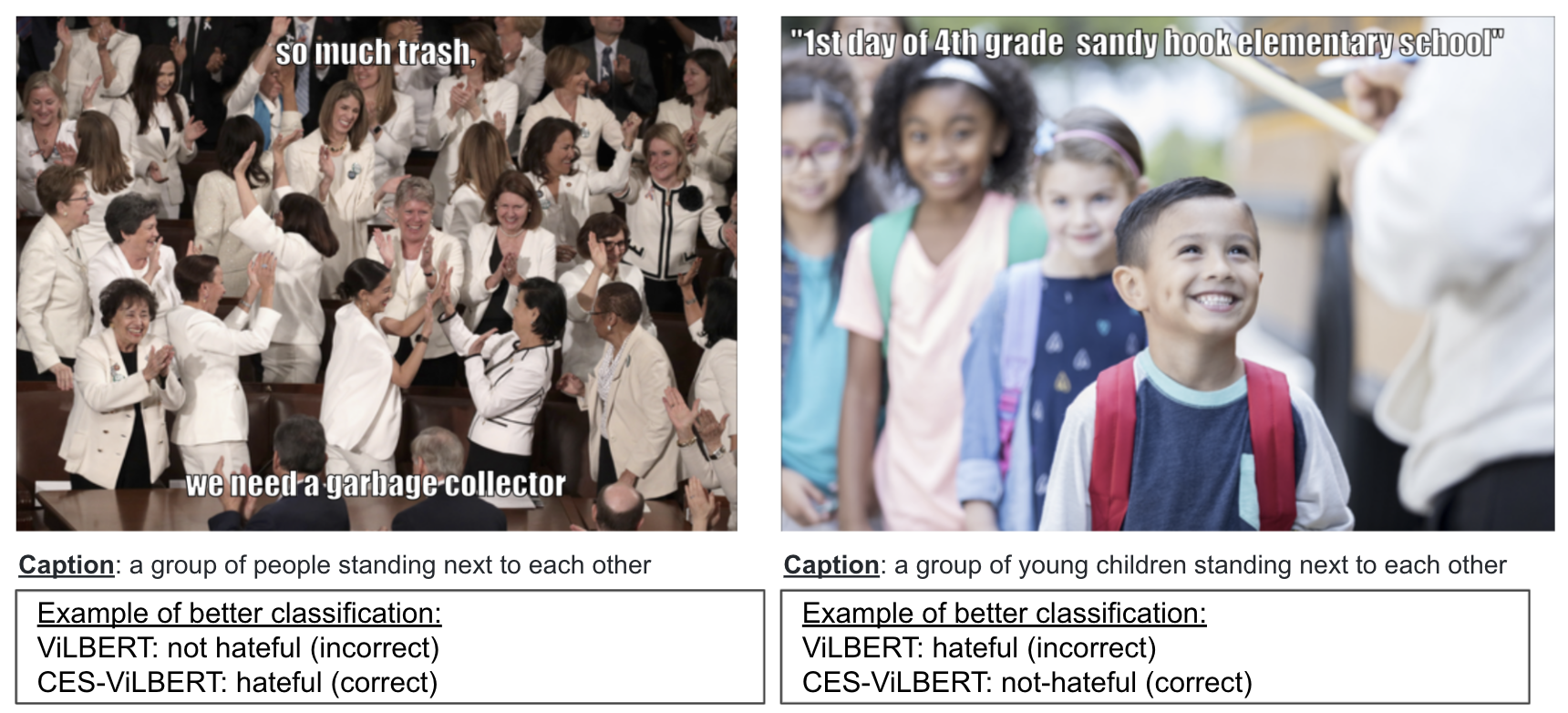}
\caption{ViLBERT results.}
\label{fig:supp_vilbert_better}
\end{figure*}

\begin{figure*}[t]
\centering
\includegraphics[width=0.8\textwidth]{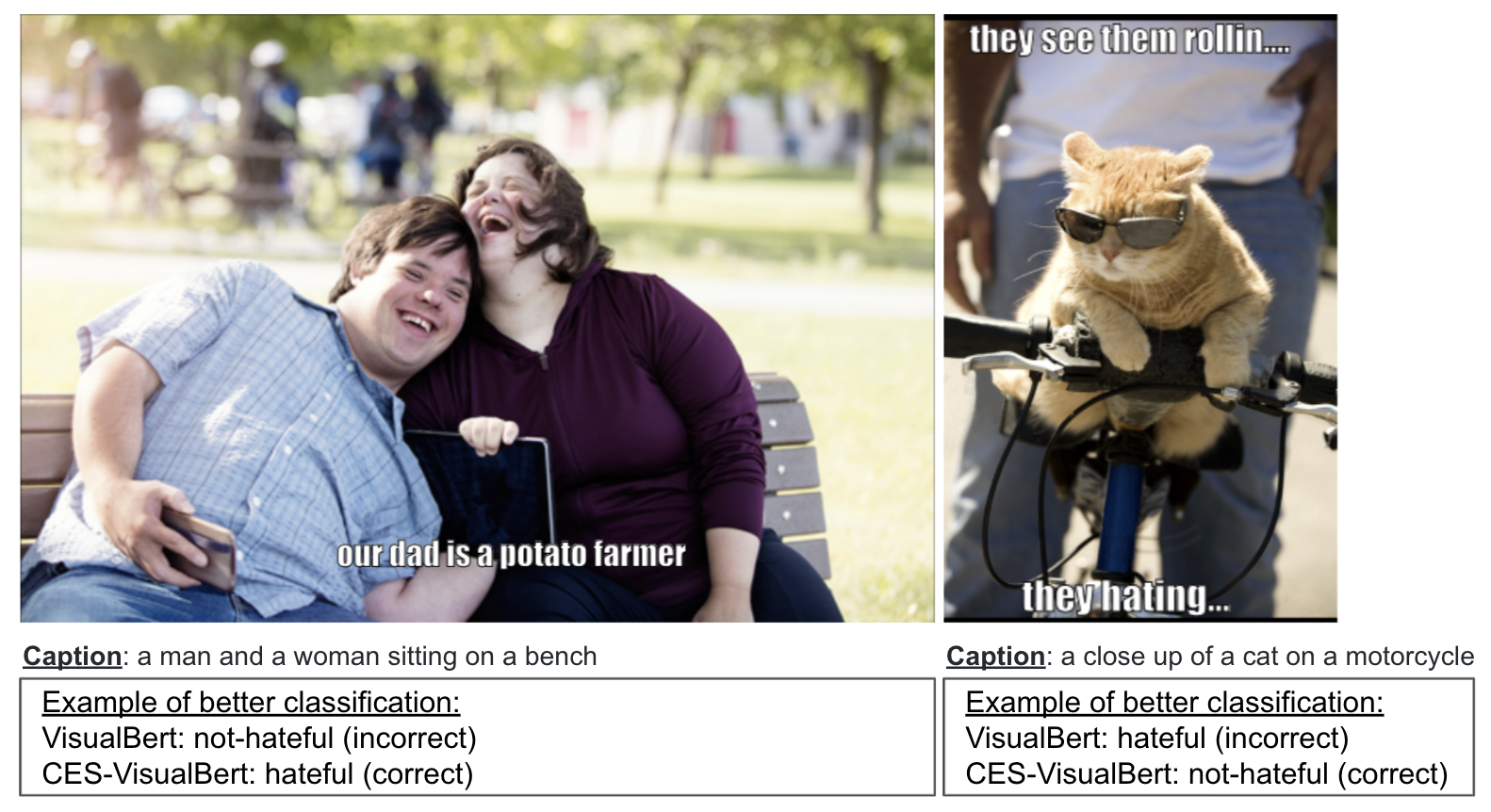}
\caption{VisualBert results.}
\label{fig:supp_visualbert_better}
\end{figure*}

Additional qualitative results for RoBERTa, BERT, ViLBERT and VisualBert are presented in Fig.\ref{fig:supp_roberta_better}, Fig.\ref{fig:supp_bert_better}, Fig.\ref{fig:supp_vilbert_better} and Fig.\ref{fig:supp_visualbert_better}, respectively.

\section{Datasets}
We use the publicly available hateful memes dataset along with Vilio's features dataset for UNITER implementation.
Locations:
\begin{itemize}
   \item Hateful Memes Dataset can be downloaded from here: \hyperlink{https://www.drivendata.org/competitions/64/hateful-memes/page/206/}{competition page}. We use Phase 1 data.
   \item Features for UNITER can be found on Kaggle: \hyperlink{https://www.kaggle.com/muennighoff/hmtsvfeats}{hmtsvfeats} and \hyperlink{https://www.kaggle.com/muennighoff/hmfeatureszipfin}{hmfeatureszipfin}
\end{itemize}
We augment the dataset with generated captions. The augmentation code can be found in the software supplementary. 

\section{Code explanation}
The code contains the following:
\begin{itemize}
  \item mmf - changes done to mmf codebase to utilize captions for BERT, RoBERTa, VisualBert, ViLBERT. 
  \item vilio\_with\_captions - changes done to Vilio codebase to support UNITER with captions.
  \item MAX-Image-Caption-Generator - provided as is for caption generations
  \item my\_hateful\_memes - code that uses MAX-Image-Caption-Generator to create captions for the dataset and the generated captions in 2 formats - CSVs for the dataset and datasets for BERT and RoBERTa's pretraining (using huggingface transformers code, not supplied)
  \item Kaggle-Ensemble-Guide - changes to the Kaggle-Ensemble-Guide codebase to generate ensembles for the results.
\end{itemize}

\section{Runtime and implementation details}
Fine-tuning models using MMF codebase: we re-used best settings as for the original Hateful memes challenge:
22000 updates for number of updates. We use weighted Adam with cosine learning rate schedule and fixed 2000 warmup steps for optimization without gradient clipping. 
Batch size: 32 for all models.
Learning rate: 5e-5 for all models.

Finetuning after pretraining for the BERT and RoBERTA models - we use 5e-6 vs 5e-5  learning rate. 

All models are trained on a single GPU GeForce GTX TITAN X with 12GB of RAM (same GPU in all runs below). Different models have different training times but mostly it took 10 to 13 hours for each model.

To continue the pre-training of the language models (BERT and RoBERTA) we used the transformers library, each pretraining was with batch size: 32 and, spread across 4 GPUs (effective batch size: 128). Pretraining runs in about 2-3 hours (100 epochs).

For the UNITER baseline, we use batch size:8 on a single GPU (learning rate: 1e-5). For ensembling it with a pre-trained RoBERTA model, we use batch size:6 with gradient accumulation: 2 (same learning rate). Each run takes roughly 1 hour. We used 36 features for the images.

\end{document}